%% file: iclr2026_conference.tex
\documentclass{article} 
\usepackage{iclr2026_conference,times}

\input{math_commands.tex}

\usepackage{algorithm}
\usepackage[noEnd,commentColor=black]{algpseudocodex}
\usepackage[most]{tcolorbox}
\usepackage{hyperref}
\usepackage{url}

\usepackage{graphicx}

\usepackage[utf8]{inputenc} 
\usepackage[T1]{fontenc}    
\usepackage{hyperref}       
\usepackage{url}            
\usepackage{booktabs}       
\usepackage{amsfonts}       
\usepackage{nicefrac}       
\usepackage{microtype}      
\usepackage{xcolor}         
\usepackage{amsmath}
\usepackage{enumitem}
\usepackage{amssymb}
\usepackage[para,online,flushleft]{threeparttable}
\usepackage[nameinlink,noabbrev]{cleveref}
\usepackage{wrapfig}

\title{Q-Learning With World Models }

\newcommand{\symfootnotetext}[2]{%
  \begingroup
  \renewcommand{\thefootnote}{#1}%
  \footnotetext{#2}%
  \endgroup
}

\author{
Perry Dong$^{*,\dagger,1}$ \quad Yueru Jia$^{*,2}$ \quad Chelsea Finn$^{1}$ \quad Dorsa Sadigh$^{1}$ \\
     \normalfont \ $^{1}$Stanford University \quad $^{2}$Peking University \\
}

\providecommand{\ours}[1][]{{\protect\color{black}{QWM\textbf{#1}}}}

\definecolor{myblue}{rgb}{1,0.5,0}
\definecolor{urlteal}{HTML}{0F766E}
\usepackage{xcolor}
\definecolor{myorange}{HTML}{E67E22}
\hypersetup{
    colorlinks = true,
    linkcolor = {myorange},
    citecolor = {myorange},
    urlcolor=myorange,
    anchorcolor = black,
}

\iclrfinalcopy 
\begin{document}

\maketitle
\symfootnotetext{*}{Equal contributions.}
\symfootnotetext{$\dagger$}{Corresponding author: \texttt{perryd@stanford.edu}}
\setcounter{footnote}{0}

\begin{abstract}

Off-policy reinforcement learning (RL) has become increasingly sample-efficient, enabling applications such as RL fine-tuning of Vision-Language-Action models into reliable, high-performing policies. World models offer a further lever for sample efficiency, as they predict state changes rather than actions alone, but their success has largely been confined to supervised policy learning. Prior model-based RL methods often optimize the policy or value function directly on imagined rollouts, which is prone to compounding bias and struggles to scale to large, high-dimensional problems such as real-world robotics, a problem that worsens with task horizon and visual complexity. In this work, we instead ask whether we can leverage world models directly on top of standard Q-learning to improve performance, while remaining trained and grounded in the real, online setting. We propose \ours{}, a framework that leverages world models to perform test-time search over imagined trajectories on top of Q-learning to select high-value actions during both online rollouts and evaluation. Since the policy and value function are trained only on real transitions, \ours{} avoids compounding model bias while still gaining the sample-efficiency benefits of predictive search. On challenging manipulation benchmarks Robomimic and LIBERO, \ours{} significantly outperforms strong prior state-of-the-art methods on both sample efficiency and performance.

\end{abstract}

\input{intro}

\input{related_work}
\input{preliminaries}

\input{method}

\input{experiments}

\input{discussion}

\bibliography{iclr2026_conference}
\bibliographystyle{iclr2026_conference}

\appendix

\clearpage

\section{Author Contributions} 

PD conceived the idea, devised the algorithm, and led the project. PD implemented the initial prototype, ran experiments, provided hands-on guidance on all experiments, and analyzed and interpreted the results. PD also wrote and positioned the paper. YJ led maintenance of the research codebase, conducted experiments, and produced the figures and videos. CF and DS contributed to the research design, advised the project,
edited and positioned the paper.

\section{Additional Experiments}
\label{ap:additional}

\begin{figure}[h]
  \centering
  \includegraphics[width=\linewidth]{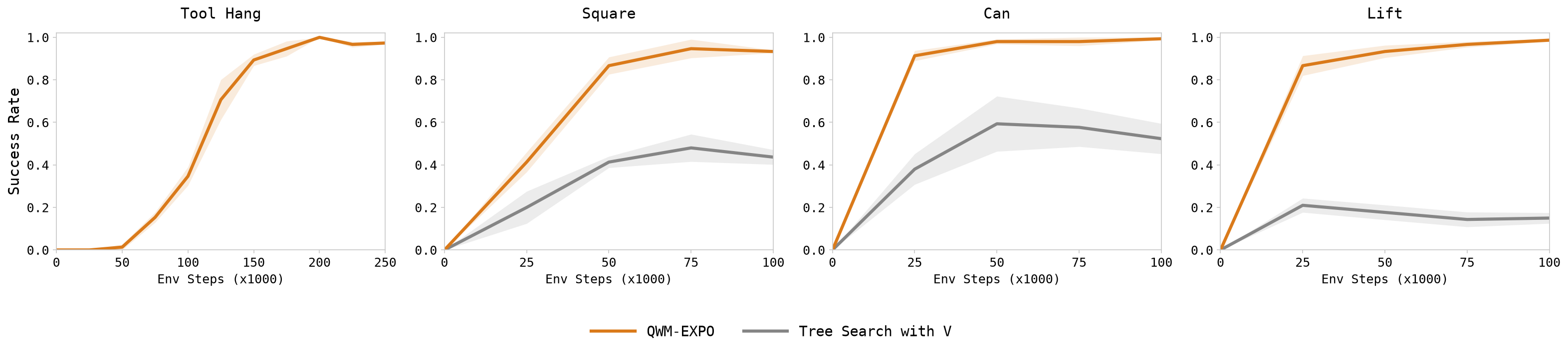}
  \caption{\footnotesize
\textbf{Search with V vs Q on top of EXPO.}
  Online success rates of \ours{}-EXPO against a search with V variant where the edit policy is learned via a state value function V. Search with Q does substantially better than the V counterpart. 
  }
  \label{fig:tree_v_expo}
\end{figure}

\begin{figure}[h]
  \centering
  \includegraphics[width=\linewidth]{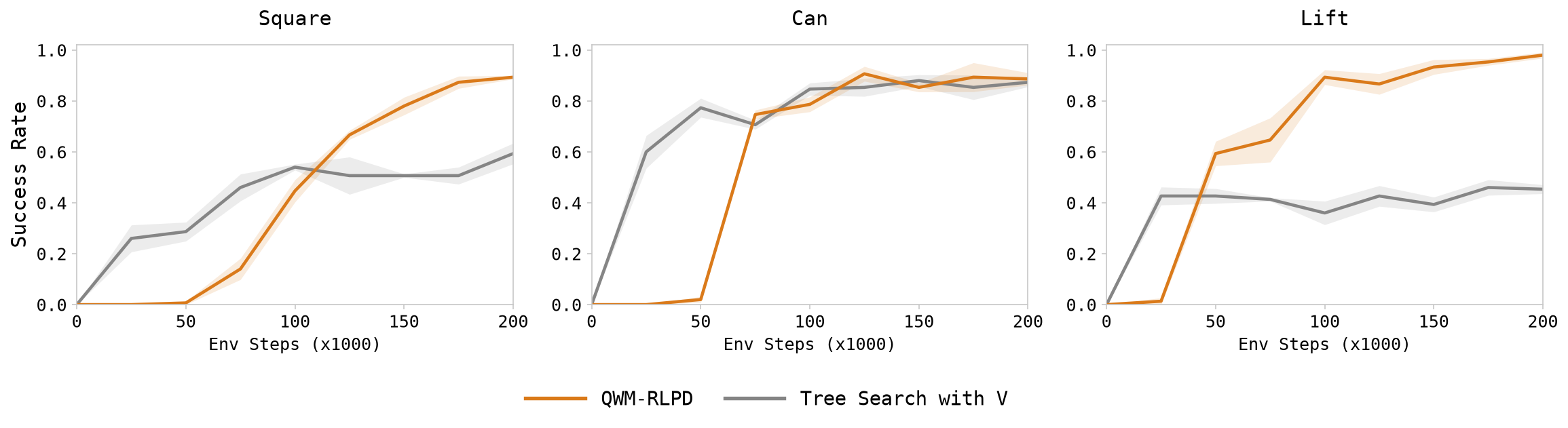}
  \caption{\footnotesize
  \textbf{Search with V vs Q as a standalone method.}
  Online success rates of \ours{}-RLPD against search with V variant with a Gaussian policy learned from V in the style of AWR. Search with Q does substantially better than the V counterpart. 
  }
  \label{fig:tree_v_rlpd}
\end{figure}

To isolate the benefit of performing test-time search with a $Q$-function, we compare
against a variant that learns only a state value function $V$ and searches with it. Here we
fit $V$, form an advantage estimate, and optimize the policy by advantage-weighted
regression~\citep{peng2019advantageweightedregressionsimplescalable}. We evaluate this variant both as the edit policy within
EXPO and as a standalone method. Results are reported in ~\Cref{fig:tree_v_expo} and ~\Cref{fig:tree_v_rlpd}. Searching with $V$ substantially underperforms searching on top of Q-learning across
all settings. We attribute this to two factors: the Q-function supplies an additional
action-conditioned estimator that $V$ alone cannot provide, which yields a stronger estimator overall, and the underlying
$Q$-learning base algorithm is itself stronger. These results motivate leveraging world models for Q-learning rather than performing search with $V$ as typically done with model-based RL.

\section{Experiment Details}

\subsection{Hyperparameters}
\label{ap:hyperparameters}

For the state-based Robomimic experiments, \ours{} and all model-free
baselines follow the same online RL protocol, including identical
demonstration sources, online interaction budgets, and optimization settings.
Unless otherwise specified, all methods start online training after 5,000
environment steps, use an offline data ratio of 0.5, do not perform additional
offline pretraining, and use an update-to-data (UTD) ratio of 20 with batch
size 256. For Q-learning-based methods, we use a critic TD discount
$\gamma=0.99$ and a target network update coefficient $\tau=0.005$.
Importantly, $\gamma$ is used only in the Bellman TD backup of the critic and
is distinct from the tree-search discount $\lambda$, which controls the
contribution of deeper future values during search. For EfficientZero V2, we
follow the official training protocol provided by the authors and use its
recommended UTD ratio of 0.4.

\begin{table}[!h]
\centering
\caption{\footnotesize
Shared optimization hyperparameters for \ours{} and model-free baselines on
Robomimic.
}
\label{tab:shared_hyperparameters}
\begin{tabular}{lc}
\toprule
Hyperparameter & Value \\
\midrule
Start training steps & 5,000 \\
Offline data ratio & 0.5 \\
Offline pretraining steps & 0 \\
UTD ratio & 20 \\
Batch size & 256 \\
Critic TD discount $\gamma$ & 0.99 \\
Target update coefficient $\tau$ & 0.005 \\
\bottomrule
\end{tabular}
\end{table}

\paragraph{\ours-specific hyperparameters.}
We instantiate \ours{} on top of two base RL algorithms, EXPO and RLPD.
For both variants, the original policy and critic optimization procedures are
kept unchanged, while world-model-guided tree search is additionally used for
action selection during online interaction and evaluation.

At each state node, the policy proposes $N$ candidate actions, and each action
is evaluated through $K$ world-model predictions. To control the size of the
search tree, we retain at most $J$ partial paths during intermediate expansion,
following the pruning procedure described in \Cref{sec:implementation}.
We use max aggregation for both intermediate-node values and leaf values.
Specifically, intermediate-node aggregation combines the values of candidate
actions and their predicted future states in the recursive value computation
in \Cref{eq:recursive-value}, while leaf aggregation combines the
$N_{\text{leaf}}$ Q-values sampled at the final search depth in
\Cref{eq:leaf-value}. The same max aggregation is used when combining
world-model branches in the root action score. The tree-search settings are summarized in
\Cref{tab:wmql_shared_search_hyperparameters}. Here, $\lambda$ is used only within the
tree-search value recursion and is independent of the critic TD discount
$\gamma$. We note that this $\lambda$ removes the factor of $\frac{1}{2}$ in \Cref{eq:root-value} and \Cref{eq:recursive-value}, and to exactly match the equations the $\lambda$ in the table should be multiplied by $\frac{1}{2}$.

\begin{table}[!h]
\centering
\caption{\footnotesize
Shared tree-search settings for state-based Robomimic experiments.
}
\label{tab:wmql_shared_search_hyperparameters}
\begin{tabular}{lc}
\toprule
Hyperparameter & Value \\
\midrule
Candidate actions $N$ & 8 \\
World-model samples per action $K$ & 8 \\
Intermediate-node aggregation & Max \\
Leaf-value aggregation & Mean \\
Search during online sampling & Yes \\
Search during evaluation & Yes \\
Nodes expanded $J$ & 1 \\
Number of leaf actions $N_{\text{leaf}}$ & 8 \\ 
Depth D & 4 \\
Value aggregation discount $\lambda$ & 0.1 \\
\bottomrule
\end{tabular}
\end{table}

\subsection{Environments}
\label{ap:environments}

We evaluate \ours{} on two widely used robotic manipulation benchmarks:
Robomimic~\citep{mandlekar2021matterslearningofflinehuman} and
LIBERO~\citep{liu2023liberobenchmarkingknowledgetransfer}. These benchmarks provide diverse manipulation
tasks with human-collected demonstrations and are commonly used to evaluate
offline-to-online robot learning algorithms.

\textbf{Robomimic.}
For state-based experiments, we evaluate \ours{} on four manipulation tasks
from Robomimic: Lift, Can, Square, and Tool Hang. These tasks cover diverse
manipulation capabilities, including object grasping and lifting, object
relocation, precise insertion, and long-horizon assembly. We use the
low-dimensional state observations provided by Robomimic, consisting of robot
proprioception and object-related features, including end-effector pose,
gripper states, and object states. The action space is a 7-DoF operational
space control (OSC) command.

Following prior work, we use different demonstration settings for different
tasks. Lift uses a 10-episode subset of the original dataset to create a more
challenging evaluation setting, Can uses the multi-human (MH) dataset, and
Square and Tool Hang use the standard proficient-human (PH) split. The PH
dataset contains demonstrations collected by a single proficient
teleoperator, while the MH dataset contains demonstrations collected by
multiple teleoperators with varying levels of expertise. All experiments
follow the standard Robomimic evaluation protocol and report online success
rates.

\textbf{LIBERO.}
For pixel-based experiments, we evaluate \ours{} on five tasks from the LIBERO
benchmark. LIBERO is a language-conditioned lifelong
robot learning benchmark designed to evaluate knowledge transfer and
generalization in robotic manipulation. It contains diverse manipulation
tasks with different object configurations, spatial arrangements, and task
semantics.

We follow the task selection and evaluation protocol used in prior work and
evaluate on five representative tasks (Task 60, Task 79, Task 29, Task 28,
and Task 2). The agent receives RGB observations from the robot cameras and
generates actions conditioned on visual observations and task instructions.
We use the standard LIBERO simulation setup and report online success rates
averaged over evaluation episodes.

\subsection{Additional World Model Details and Prediction Quality}
\label{ap:world_model}

\begin{figure}[t]
    \centering
    \includegraphics[width=\linewidth]{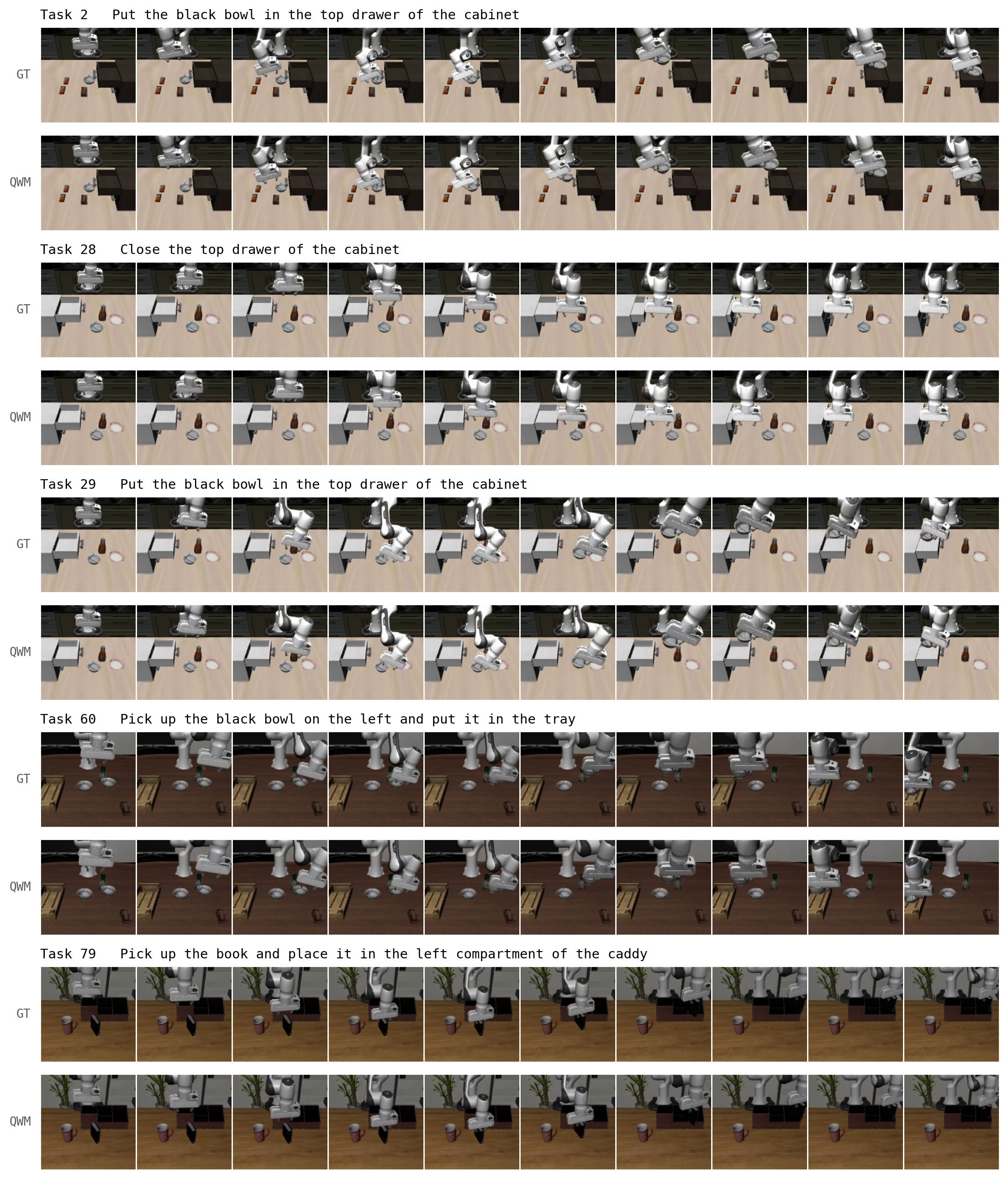}
    \caption{\footnotesize
    \textbf{Qualitative comparison of world-model predictions on LIBERO.}
    We compare the current observation, ground-truth future observations, and
    action-conditioned predictions over short rollout horizons. The predicted
    observations capture the major robot and object changes relevant to
    tree-search action evaluation, while prediction errors become more visible
    at later rollout steps.
    }
    \label{fig:wm_predictions}
\end{figure}

We provide additional training and inference details for the world models used
in our experiments. For the state-based experiments, the dynamics model is
pretrained offline on demonstration transitions $(s,a,s')$ using an MSE
objective with Adam, a learning rate of $3\times10^{-4}$, batch size 256, and
100k gradient steps, and is kept fixed throughout subsequent policy and critic
optimization. For the vision-based experiments, we fine-tune the
action-conditioned Wan2.2-TI2V-5B world model on demonstration videos paired
with aligned raw action sequences. Before training, video clips are encoded
once by the frozen Wan2.2 VAE and text prompts are encoded by the frozen
umT5-XXL encoder, allowing both representations to be cached and removed from
the training loop. We then jointly fine-tune the diffusion transformer and
action encoder using the standard Wan2.2 flow-matching objective, while keeping
the VAE and text encoder frozen. Training uses AdamW with a learning rate of
$1\times10^{-5}$, a per-GPU batch size of 1 on 8 GPUs, bfloat16 precision,
gradient checkpointing, and 200k gradient steps. For LIBERO, we train on 5-frame $128\times128$ clips and model the
agent-view and wrist-camera observations as two video streams. During tree
search, however, each world-model query advances the rollout by only one
step: we use the second frame of the generated clip as the predicted next
observation, and then condition a new generation on this prediction for
subsequent tree expansion. At inference time, we use 1 denoising
step. To examine the
next-step generation quality of the resulting world model,
\Cref{fig:wm_predictions} compares the generated next observation with the
corresponding ground-truth next observation conditioned on the same current
observation and action. The predictions capture the task-relevant changes in
robot configuration and object motion at the next step, supporting their use
as predicted future observations during tree search.

\subsection{Comparisons}
\label{ap:comparisons}

We evaluate \ours{} against both state-of-the-art model-free algorithms and model-based algorithms, as while \ours{} uses a learned model, it does not use data from the model for training and is trained on top of standard Q-learning. 

\textbf{Model-free baselines.}
The first set of comparisons is against state-of-the-art model-free algorithms. As \ours{} learns with only data in the real world instead of from the model, we compare against other such methods to evaluate effectiveness.

\textbf{EXPO}~\citep{dong2025expostable}.
EXPO jointly learns an expressive base policy and a lightweight edit policy, where the edit policy transforms actions sampled from the base policy toward higher-value distributions.

\textbf{IDQL}~\citep{hansenestruch2023idqlimplicitqlearningactorcritic}.
IDQL trains a diffusion policy through behavior modeling and performs implicit policy extraction using best-of-$N$ sampling, selecting the candidate action with the highest Q-value.

\textbf{RLPD}~\citep{ball2023efficient}.
RLPD is a sample-efficient off-policy RL method that leverages replay data with high update-to-data ratios, critic ensembles, and a Gaussian policy.

\textbf{QSM}~\citep{pmlr-v235-psenka24a}.
QSM incorporates gradients from the learned Q-function into the diffusion training objective, guiding the denoising process toward high-value actions through score matching.

\textbf{DSRL}~\citep{wagenmaker2025steeringdiffusionpolicylatent}.
DSRL adapts a frozen diffusion-based behavior cloning policy by performing reinforcement learning over the initial noise used for action generation, learning a policy that predicts the noise seed.

\textbf{QAM}~\citep{li2026q}.
QAM uses adjoint matching to propagate value gradients through the diffusion trajectory and constructs policy optimization objectives based on the learned Q-function.

\textbf{FQL}~\citep{park2025flowqlearning}.
FQL trains a one-step flow policy to maximize Q-values learned with a temporal-difference objective, while regularizing the policy toward a behavior-cloned flow policy.

\textbf{Model-based baselines.} The second category of comparison methods consists of model-based RL algorithms, which similarly learn a dynamics model but, unlike our approach, use this model directly for training. 

\textbf{TD-MPC2}~\citep{hansen2024td}.
TD-MPC2 learns an implicit latent world model and performs short-horizon MPPI planning in latent space. It evaluates sampled action sequences using predicted rewards and bootstrapped terminal values, with a learned maximum-entropy policy prior guiding the sampling process.

\textbf{EfficientZero V2}~\citep{pmlr-v235-wang24at}.
EfficientZero V2 learns a MuZero-style latent world model and performs Sampling-based Gumbel tree search for both discrete and continuous control. The search provides bootstrapped policy and value targets that are reused for policy and value training.

\end{document}

%% file: math_commands.tex
\usepackage{amsmath,amsfonts,bm}

\def\eqref#1{equation~\ref{#1}}

\def\1{\bm{1}}

\DeclareMathAlphabet{\mathsfit}{\encodingdefault}{\sfdefault}{m}{sl}
\SetMathAlphabet{\mathsfit}{bold}{\encodingdefault}{\sfdefault}{bx}{n}



%% file: intro.tex
\section{Introduction} \label{sec:introduction}


Recent advances in off-policy reinforcement learning (RL) have substantially improved its sample efficiency, enabling applications such as RL fine-tuning of Vision-Language-Action (VLA) models into highly reliable, high-performing policies~\citep{intelligence2025pi06vlalearnsexperience,dong2026expoftsampleefficientreinforcementlearning}. As RL is applied to increasingly complex scenarios, further gains in sample efficiency become increasingly valuable, if not essential. World models offer a natural lever for closing this gap, as they have the ability to predict changes in state rather than actions alone. So far, however, their success has been largely confined to supervised policy learning. This raises a natural question: is there a simple way to leverage world models to improve the performance of online RL, without resorting to traditional model-based RL approaches that are prone to compounding the world model's bias into the training process?


Prior work on model-based RL has explored learning a dynamics model and optimizing a policy or value function directly within it~\citep{hansen2024td,hafner2024masteringdiversedomainsworld}, but scaling this approach to large, high-dimensional problems such as real-world robotics remains difficult: policies trained primarily on rollouts inside an imperfect world model inherit and compound its biases, and this problem worsens as task horizon and visual complexity increase. Rather than using the world model as a substitute for real interaction, we ask whether it can instead sharpen an RL agent that remains trained and grounded in the real, online setting. In this work, we study Q-learning with a world model, where the model serves not as a replacement for environment interaction but as a way for improving performance.

Our key insight is that a learned world model can be used on top of standard Q-learning to enable test-time search over actions by imagining future trajectories, and that this improved action selection alone directly translates into higher performance for Q-learning methods. We propose Q-Learning with World Models (\ours{}), a framework that leverages world models for test-time search directly on top of standard Q-learning. Instead of using the state value function to conduct search as in typically done in model-based RL, the learned Q-function also enables more powerful search to evaluate candidate actions and select the highest-value one before it is executed. This test-time search over actions is applied both during online rollouts, to obtain better data for online RL, and at evaluation time, to act more effectively with the learned policy. Rather than sampling a single action from the policy or sampling multiple actions and selecting the one with the highest Q-value, \ours{} leverages the imagination capability of world models to select the best action for downstream performance, while avoiding the compounding bias and instability of policies trained inside a learned model.


Our main contribution is \ours{}, a simple and general framework for leveraging learned world models to improve Q-learning through test-time search. Because the underlying policy and value function are still trained online against real environment transitions, \ours{} avoids the bias accumulation that afflicts methods relying primarily on imagined rollouts, while still reaping the sample-efficiency benefits such a predictive model can offer at decision time. We evaluate \ours{} on challenging manipulation benchmarks, Robomimic and LIBERO, across both state-based and visual observation settings, and find that it significantly outperforms strong prior methods on both sample efficiency and performance.

\begin{figure}[t]
\centering
\includegraphics[width=\linewidth]{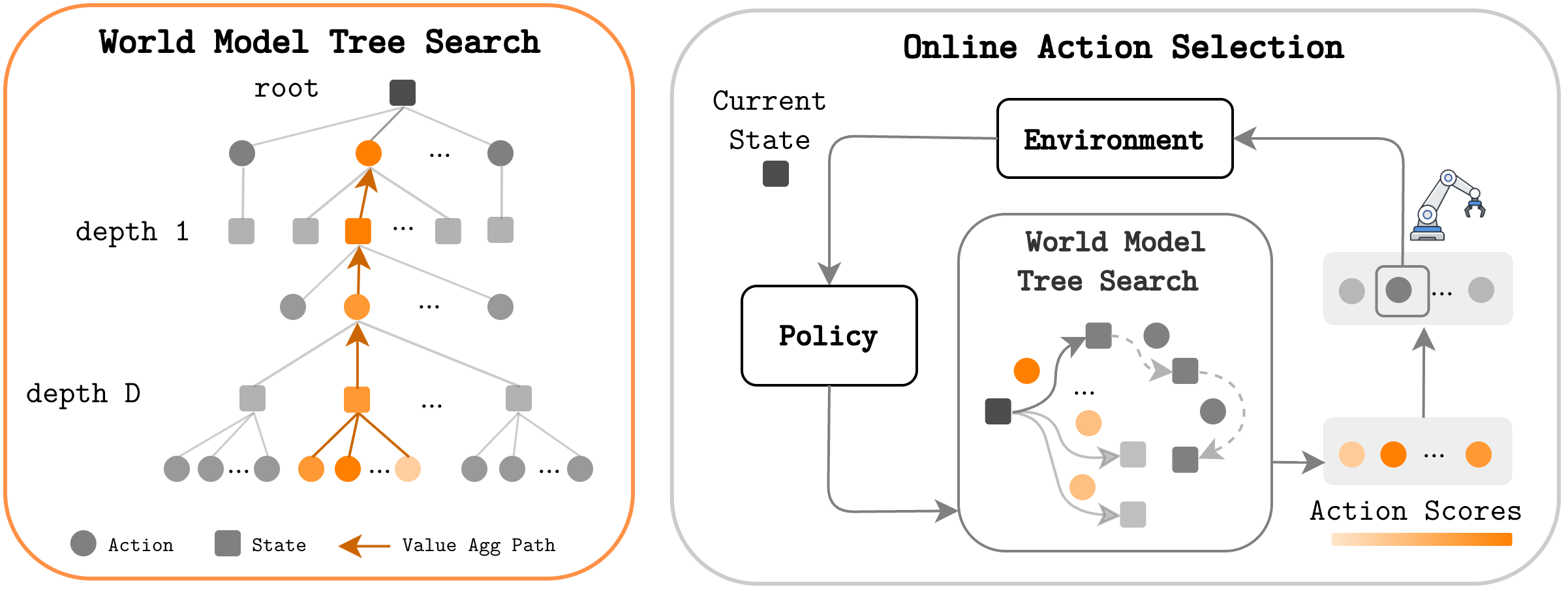}
\caption{
\textbf{Overview of \ours{}.}
\textbf{Left: World-model tree search.}
From the current state, the policy proposes candidate actions and the world model imagines future states. The value of each initial action is computed by aggregating value of intermediate and leaf nodes.
\textbf{Right: Online action selection.}
At each environment step, \ours{} uses tree-search scores to select among policy-proposed actions before executing the selected action in the environment. Tree search is used for both online sampling and evaluation, while the policy and critic are trained only on real environment transitions.
}
\label{fig:tree_search}
\end{figure}

%% file: related_work.tex
\section{Related Work} \label{sec:related_work}

\textbf{Reinforcement learning with prior data. } To improve performance and sample efficiency of online RL, prior works have studied the problem of using an offline dataset to accelerate online learning. Popular approaches involve balancing exploration and exploitation~\citep{yang2023hybrid,zhang2023policyexpansion,mark2023offline}, calibrating value estimates~\citep{nakamoto2024calibrate,dong2026reallyneedpretrainqfunctions}, or retaining offline data alongside newly collected online data during fine-tuning~\citep{vecerik2018leveragingdemonstrations,nair2021awacacceleratingonline,ball2023efficient,dong2025reinforcementlearning,dong2025expostable}. Offline RL algorithms have also been applied directly to online fine-tuning~\citep{fujimoto2019offpolicydeepreinforcementlearning,fujimoto2021minimalistapproachofflinereinforcement,hansenestruch2023idqlimplicitqlearningactorcritic,park2025flowqlearning,dong2025mattersbatchonlinereinforcement,dong2026tqlscalingqfunctionstransformers,dong2026valueflows}, where the prior data is first trained with offline RL before obtaining new samples online. Our work differs from these lines of research in that we focus on leveraging prior data through world models to improve online RL, rather than on how prior data is incorporated into the RL objective itself; this makes our approach applicable to any RL fine-tuning algorithm. 
 
\textbf{Model-based RL. } Model-based RL methods learn a dynamics model of the environment and use it for policy or value learning. One family of methods generates additional synthetic transitions by rolling the learned model forward and treats them as if they were real experience~\citep{sutton1991dyna, gu2016continuousdeepqlearningmodelbased,pmlr-v78-kalweit17a,kurutach2018modelensembletrustregionpolicyoptimization,kaiser2024modelbasedreinforcementlearningatari}, or otherwise optimizes the policy directly against imagined trajectories~\citep{zhang2019solardeepstructuredrepresentations,hafner2020dreamcontrollearningbehaviors,hafner2022masteringataridiscreteworld,hafner2024masteringdiversedomainsworld,hafner2025trainingagentsinsidescalable} or through short branched rollouts mixed with real data~\citep{feinberg2018modelbasedvalueestimationefficient,janner2021trustmodelmodelbasedpolicy}. A second family instead uses the learned model only to plan over imagined rollouts with trajectory optimizers such as the cross-entropy method or MPPI~\citep{nagabandi2018neural,ebert2018visualforesightmodelbaseddeep,chua2018,hansen2022temporaldifferencelearningmodel,hansen2024td}; some of these methods further learn a terminal value function to bound the planning horizon~\citep{hansen2022temporaldifferencelearningmodel,hansen2024td,kim2026cosmospolicyfinetuningvideo}. A third family combines a learned (or known) model with Monte Carlo tree search guided by a learned value function, mainly for discrete-action, densely-simulated domains~\citep{silver2017masteringchessshogiselfplay,Schrittwieser_2020,ye2021masteringatarigameslimited,schrittwieser2021onlineofflinereinforcementlearning}, with recent works extending it to continuous control~\citep{hubert2021learningplanningcomplexaction,pmlr-v235-wang24at}. These methods rely on some combination of having a known dynamics model, discrete actions, or bootstrapping policy and value targets from search statistics computed over model-imagined rollouts. Furthermore, many of these works rely on abundant simulated data, where model error and sample budget are far less of a concern than in real-world, sparse-reward robotic manipulation. Across all of these approaches, the policy or value function trained substantially on model-generated rollouts inherits and compounds the model's own biases directly into training, an issue that grows with task horizon and visual complexity~\citep{janner2021trustmodelmodelbasedpolicy,chua2018,kaiser2024modelbasedreinforcementlearningatari,hafner2020dreamcontrollearningbehaviors}. \ours{} sidesteps this issue by leveraging the model purely at test-time rather than a source of additional training data on top of standard Q-learning, and train the policy and critic on real data.

\paragraph{World models for VLAs. }
A closely related line of work applies world models to VLA policies. VLAW~\citep{guo2026vlawiterativecoimprovementvisionlanguageaction} and World-VLA-Loop~\citep{liu2026worldvlaloopclosedlooplearningvideo} alternate between fine-tuning world models on real rollouts and using it to generate synthetic training data for policy learning. WMPO~\citep{zhu2025wmpoworldmodelbasedpolicy} and World4RL~\citep{jiang2026world4rldiffusionworldmodels} run on-policy RL directly on imagined rollouts, while WoVR~\citep{jiang2026wovrworldmodelsreliable} targets the resulting reward hallucination and OOD drift via keyframe-initialized rollouts and policy-aligned co-evolution. RISE~\citep{yang2026riseselfimprovingrobotpolicy}, World-Gymnast~\citep{sharma2026worldgymnasttrainingrobotsreinforcement}, and GigaBrain-0.5M
~\citep{gigabrainteam2026gigabrain05mvlalearnsworld} score imagined rollouts with a learned progress/value model or VLM and use the resulting rewards for policy-gradient updates. All of these use the world model primarily as a training data generator, which is prone to inheriting compounding bias. SAILOR~\citep{jain2025smoothseaskilledsailor} learns a world model and reward model to guide imitation and recover from out-of-support states. \ours{} can be applied to any RL fine-tuning methods for VLAs with a Q-function, and instead of using the world model to generate data or guide imitation learning, we leverage world models to improve Q-learning at test time. 
 
\paragraph{Test-time scaling and best-of-$N$ inference.}

A substantial body of work has demonstrated that test-time scaling, for example via best-of-$N$ sampling, can significantly improve the performance of generative models across domains, including autoregressive language models \citep{wang2022selfconsistency,snell2024scalingllm,chow2024inferenceaware}, diffusion models \citep{ma2025itsdiffusion}, and policy learning \citep{dong2025expostable,hansenestruch2023idqlimplicitqlearningactorcritic,dong2026fastervalueguidedsamplingfast,chen2023offline}. In most of these settings, candidates are scored independently by a verifier or value function, and only the single best sample is retained, without modeling how a candidate's near-term consequences unfold. \ours{} can instead be viewed as a form of test-time scaling in which search evaluates candidates on their predicted downstream value, yielding a substantially better estimate of which actions are best.

%% file: preliminaries.tex
\section{Preliminaries} \label{sec:preliminaries}

We consider a Markov decision process (MDP) specified by the tuple $\{\mathcal{S}, \mathcal{A}, \rho, r, \gamma, T\}$, where $\mathcal{S}$ denotes the state space, $\mathcal{A}$ denotes the action space, $\rho(s)$ is the distribution over initial states, $r:\mathcal{S}\times\mathcal{A}\rightarrow\mathbb{R}$ is the reward function, $\gamma\in[0,1)$ is the discount factor, and $T(s'|s,a)$ is the transition probabilities governing the environment. The goal of RL is to find a policy $\pi$ that maximizes the expected discounted return $\mathbb{E}_{\pi}\left[\sum_{t=0}^{T}\gamma^t r(s_t, a_t)\right]$.

In this paper, we study how a learned world model can be used to improve online RL fine-tuning at decision time. Concretely, we assume access to a dataset $\mathcal{D}_{\text{offline}} = \{(s,a,r,s')\}$ of environment transitions -- either collected offline or accumulated online -- which is used to pretrain a world model $M_\psi(s'|s,a)$ that predicts the next state given the current state and action. The world model is used at decision time. During online training, the agent collects tuples $(s_t,a_t,r_t,s_{t+1})$ through environment interaction; these are appended to a replay buffer $\mathcal{D}$ and used to update the policy and Q-function toward higher returns. Our central question is how to leverage the world model to improve the performance of Q-learning. 

We focus on off-policy RL, where the Q-function estimates the discounted return of the policy given a state and action, and is trained with TD learning:
\begin{equation}
    \mathcal{L}(\phi) = \mathbb{E}_{(s_t, a_{t}, s_{t+1}) \sim \mathcal{D}}\left[\left(r_t + \gamma Q_{\phi'}(s_{t+1}, \tilde{a}^*_{t+1}) - Q_\phi(s_t, a_{t})\right)^2\right],
\end{equation}
where $Q_{\phi'}$ is a target network and $\tilde{a}^*_{t+1}$ is the next action selected by the RL policy. We implement \ours{} on top of two state-of-the-art off-policy Q-learning algorithms.

\textbf{EXPO. } The first is EXPO~\citep{dong2025expostable}, a high-performing online RL method that couples a base policy with a lightweight edit policy. EXPO maintains two parameterized policies: a base flow policy---in our case, the base policy $\pi_\text{base}$, trained with a supervised loss---and an edit policy $\pi_\text{edit}$, trained to maximize the Q-function:
{\small
\begin{equation}
\begin{split}
    \mathcal{L}(\pi_\text{edit}) = -\mathbb{E}_{(s_t,a_{t})\sim\mathcal{D},\;\hat{a}_{t}\sim\pi_\text{edit}}
    \bigl[Q_\phi(s_t,\, a_{t} + \hat{a}_{t}) - \alpha \log \pi_\text{edit}(\hat{a}_{t}\mid s_t, a_{t})\bigr].
\end{split}
\end{equation}
}The edit policy predicts an edit $\hat{a}_t$, clipped to $[-\beta, \beta]$, that is added to the base action $a_t$ to produce an edited action $\tilde{a}_t = a_t + \hat{a}_t$. This design avoids backpropagating gradients through the base policy while still grounding TD updates in near-optimal actions. The final inference and TD-backup policy is an on-the-fly (OTF) policy that selects the value-maximizing candidate among $N$ sampled base and edited actions:
\begin{equation}
    \tilde{a}^* = \underset{a \;\in\; \bigcup_{i=1}^{N}\{a_i,\,\tilde{a}_i\}}{\arg\max}\; Q_\phi(s, a),
\end{equation}
where $Q_\phi$ is trained with a standard TD objective.

\textbf{RLPD.}
The second is RLPD~\citep{ball2023efficient}, a sample-efficient off-policy RL algorithm that combines offline and online replay data with high update-to-data (UTD) ratios, an ensemble of critics, and a Gaussian actor. RLPD trains the actor $\pi_\theta$ with a SAC-style~\citep{haarnoja2018softactorcriticoffpolicymaximum} entropy-regularized objective,
\begin{equation}
    \mathcal{L}(\pi_\theta) = -\mathbb{E}_{s_t \sim \mathcal{D},\; a_t \sim \pi_\theta}\bigl[Q_\phi(s_t, a_t) - \alpha \log \pi_\theta(a_t \mid s_t)\bigr],
\end{equation}
and stabilizes the high-UTD critic updates using an ensemble of $M$ critics with random subsets used for the target computation, together with layer normalization. At each update, transitions are drawn via symmetric sampling, with equal proportions from the offline buffer and the online replay buffer.

%% file: method.tex
\section{Method} \label{sec:method}

In this section, we present the key component of \ours{} to leverage world models to perform test-time search for better performance for Q-Learning. We instantiate \ours{} as a tree search over actions on top of standard Q-learning We first describe the approach for tree search with world models, split into three parts: constructing the search tree (\Cref{sec:tree-search}), aggregating the value of expanded nodes into a value for each proposed action (\Cref{sec:sampling-eval}), and searching with the Q-function (\Cref{sec:large}); we then discuss implementation details (\Cref{sec:implementation}).

\subsection{Tree Search With the World Model}
\label{sec:tree-search}

Our goal is to construct a tree such that searching over it yields good actions and adequate coverage of possible futures, and is general enough to be compatible with different problem settings. We construct the tree governed by three quantities: the number of actions sampled at each state $N$, the number of next states sampled per action $K$, and the search depth $D$.

Concretely, at each decision step, we set the root of the tree to the current state $s_0$ where depth $0$ represents the root of the tree. Nodes alternate between state layers and action layers, with one state layer followed by one action layer constituting a single level of depth. At a state node at level $d$, the tree is expanded by drawing $N$ candidate actions from the policy, $\{a_d^{n}\}_{n=1}^N \sim \pi_\theta(\cdot \mid s_d)$. At each action node, the tree is expanded and form depth $d+1$ by querying the world model $K$ times to obtain $K$ predicted future states, $\{s^{k}_{d+1}\}_{k=1}^{K} \sim M_\psi(s_d, a_d^{n})$.

\begin{figure}[t!]
\centering
\includegraphics[width=\columnwidth]{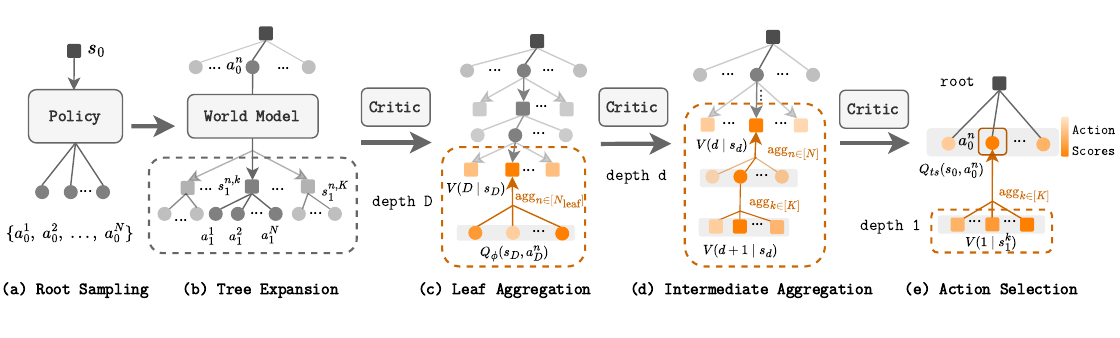}
\caption{
\small
\textbf{Tree Search and Value Aggregation in \ours{}.}
From the current state $s_0$, the policy proposes $N$ candidate actions.
The world model predicts the corresponding future states, and this expansion
repeats to depth $D$. Rather than evaluating only the leaf nodes, values are
aggregated from intermediate nodes, where deeper imagined branches are
discounted according to the tree-search discount $\lambda$.
}
\label{fig:tree-diagram}
\end{figure}

We repeat this expansion recursively to depth $D$ (\Cref{fig:tree-diagram}). At the final layer, $N_{\text{leaf}}$ actions are sampled from the policy. The tree is intended to give a short-horizon prediction of the consequences of an action, not to plan exhaustively -- this keeps rollout lengths short enough to prevent the world model's own prediction error to accumulate.

\subsection{Aggregating the Value}
\label{sec:sampling-eval}

Rather than scoring only the leaves and propagating value to the initial
action, we observe that the value of every state-action pair in the
tree is useful for action selection. The value of a node can either be expressed as a function of state-action value $Q_{\phi}$ or a function of state value $V_{\phi}$, each with complementary trade-offs. As such, we express the value of a node as a
combination of these two estimators of the same underlying quantity.

\textbf{State-action value estimator.} At depth $d$, we can score each of the $N$ sampled actions
$a_d^{n} \sim \pi_\theta(\cdot \mid s_d)$ directly with the critic,
which already estimates the return of the full remaining trajectory:
\begin{equation}
    V_Q\big(d \mid s_d\big) \;=\; \operatorname*{agg}_{n \in [N]}\; Q_\phi\big(s_d,\, a_d^{n}\big).
    \label{eq:v-bootstrap}
\end{equation}
This estimator is low-variance and independent of the remaining search depth and error from the model,
but relies entirely on $Q_\phi$ being learned accurately.

\textbf{State value estimator.} We can alternatively estimate using the world model's predicted per-step reward
$r_\psi(s_d, a_d^n)$ and a discounted value of the next state:
\begin{equation}
    V_r\big(d \mid s_d\big) \;=\; \operatorname*{agg}_{n \in [N]}\left[\; \operatorname*{agg}_{k \in [K]}\Big[\, r_\psi\big(s_d, a_d^{n}\big) \;+\; \lambda\, V\big(d+1 \mid s_{d+1}^{n,k}\big)\Big]\right],
    \label{eq:v-rollout}
\end{equation}
where $s_{d+1}^{n,k} = M_\psi\big(s_d, a_d^{n}\big)$ and $r_\psi(s_d, a_d^n)$ is the learned reward model. At the final layer, $N_{\text{leaf}}$ actions are sampled from the
policy and there is no rollout left to recurse into, so the leaf value
reduces to state value which can be computed as an aggregate over Q-values for actions in that state,
\begin{equation}
    V\big(D \mid s_D\big) \;=\; \operatorname*{agg}_{n \in [N_{\text{leaf}}]}\; Q_\phi\big(s_D,\, a_D^{n}\big).
    \label{eq:leaf-value}
\end{equation}
The state value estimator exploits the full imagined rollout but compounds world-model error with depth.

\textbf{Combined node value.} The two estimators have complementary failure
modes: $V_Q$ is depth-independent but blind to the rollout, while $V_r$ uses
the rollout but accumulates simulation error. We take advantage of both estimators to get a combined node value:
\begin{equation}
    V\big(d \mid s_d\big) \;=\;  \tfrac{1}{2} \Big(V_Q\big(d \mid s_d\big) +
V_r\big(d \mid s_d\big) \Big)
    \label{eq:recursive-value}
\end{equation}
which can be extended to a weighted average $\alpha V_Q + (1-\alpha)
V_r$, where $\alpha \in [0,1]$.

\paragraph{Root value.} The value of each originally sampled root action
$a_0^n$ follows the same computation at $d=0$:
\begin{equation}
    Q_{ts}(s_0, a_0^n) \;=\; \tfrac{1}{2}\left(Q_\phi(s_0, a_0^n) \;+\; \Big[r_\psi(s_0, a_0^n) + \operatorname*{agg}_{k \in [K]}\lambda\, V\big(1 \mid s_1^{k}\big)\Big]\right), \qquad s_1^{k} = M_\psi(s_0, a_0^n).
    \label{eq:root-value}
\end{equation}
The action ultimately selected is a max or softmax over the original sampled
actions, weighted by their tree-search values $Q_{ts}(s_0, a_0^n)$. We use
this search procedure in two places: to select actions during training, and
to act at evaluation time.

\subsection{Searching with the Q-function. }
\label{sec:large}
Because the tree grows very large very quickly with depth, due to computational constraints we in practice design a heuristic to select only a subset of nodes to expand and prune the rest using the Q-function as a search policy. Let $J$ be the number of paths that will be expanded. Let $\mathcal{B}_d$ denote the surviving set of $J$ partial paths after pruning at depth $d$ (with $\mathcal{B}_0 = \{a_0^n\}^{N}_{n=1}$, the single root path under consideration). Each surviving path $b \in \mathcal{B}_d$ carries an accumulated discounted score
\begin{equation}
    \Sigma(b, d) \;=\; \sum_{d'=1}^{d} \lambda^{d'}\, Q_\phi\big(s_{d'}^{b}, a_{d'}^{b}\big),
    \label{eq:beam-accum}
\end{equation}
where $\big(s_{d'}^{b}, a_{d'}^{b}\big)$ is the state-action pair visited by path $b$ at depth $d'$.

At depth $d+1$, every surviving path $b \in \mathcal{B}_d$ is expanded into $N$ candidate children $\{a_{d+1}^{b,n}\}_{n=1}^{N} \sim \pi_\theta(\cdot \mid s_{d+1}^{b})$, each scored by its prospective cumulative sum:
\begin{equation}
    \tilde\Sigma(b, n, d+1) \;=\; \Sigma(b, d) \;+\; \lambda^{d+1}\, Q_\phi\big(s_{d+1}^{b,n}, a_{d+1}^{b,n}\big).
    \label{eq:beam-candidate-score}
\end{equation}
The surviving set at depth $d+1$ is the top $J$ over \emph{all} candidates, ranked by \Cref{eq:beam-candidate-score}:
\begin{equation}
    \mathcal{B}_{d+1} \;=\; \operatorname*{top\text{-}}J \Big\{ \tilde\Sigma(b,n,d+1) \;:\; b \in \mathcal{B}_d,\; n \in [N] \Big\}.
    \label{eq:beam-prune}
\end{equation}
At depth $D$, each surviving path $b \in \mathcal{B}_{D-1}$ samples $N_{\text{leaf}}$ final actions, whose values are aggregated per-beam:
\begin{equation}
    L(b) \;=\; \operatorname*{agg}_{n \in [N_{\text{leaf}}]}\; Q_\phi\big(s_D^{b}, a_D^{b,n}\big).
    \label{eq:beam-leaf}
\end{equation}
The root's tree-search value combines all surviving beams. Intuitively, \Cref{eq:beam-prune} selects survivors by the \emph{additive discounted sum} $\tilde\Sigma$. When $J$ equals the max number of nodes, no candidates are discarded at \Cref{eq:beam-prune} ($\mathcal{B}_d$ always contains all $N^d$ paths). When $J$ is less than the max number of nodes, \Cref{eq:beam-prune} discards candidates at every intermediate depth based on \Cref{eq:beam-candidate-score}.

\subsection{Practical Implementation}
\label{sec:implementation}

We instantiate the action-conditioned world model in two forms depending on
the observation modality. In practice, because we operate in sparse reward settings where rewards are zero except for the terminal state, we do not learn a reward model on top of the world model. For low-dimensional state observations, we use a
deterministic residual dynamics model
$M_\psi(s_t,a_t)=s_t+\Delta_\psi(s_t,a_t)$ implemented as a three-layer MLP
with hidden dimension 256. The model takes the Robomimic low-dimensional state
representation, including end-effector pose, gripper states, and object
features, together with the 7-DoF action as input. It is pretrained offline on
demonstration transitions using an MSE objective.
For pixel observations, we adapt Wan2.2-TI2V-5B~\citep{wan2025} into an action-conditioned video
world model by introducing an MLP-based action encoder. Specifically, each
robot action is mapped into action tokens through a three-layer MLP, which are
then concatenated with the text conditioning tokens and provided to the
diffusion transformer as additional conditioning inputs. The action encoder and
diffusion transformer are jointly fine-tuned while keeping the VAE and text
encoder frozen. The model is trained on demonstration video clips with aligned
action sequences using the standard Wan2.2 flow-matching objective. During
inference, the world model predicts short-horizon future observations
conditioned on candidate actions. For LIBERO, we generate 5-frame $128\times128$ video clips, and use the predicted next frame as the subsequent observation for iterative rollout and tree search.




%% file: experiments.tex
\section{Experiments} \label{sec:experiment} 

The goal of our experiments is to answer the following core questions:

\begin{enumerate}[start=1,label={(\bfseries Q\arabic*)}]
    \item How does \ours{} perform compared to state-of-the-art model-free and model-based RL methods?
    \item How does \ours{} perform compared to the base method it is implemented on top of?
    \item Does \ours{} scale to a pixel-based setting? 
    \item What components of \ours{} are most important for performance? 
\end{enumerate}

\textbf{Environments.}
We evaluate \ours{} on challenging robotic manipulation tasks from Robomimic~\citep{mandlekar2021matterslearningofflinehuman} and
LIBERO~\citep{liu2023liberobenchmarkingknowledgetransfer}, where a 7-DoF robot arm is required to complete diverse manipulation behaviors under sparse task-completion rewards.
For Robomimic, we consider four tasks: Lift, Can, Square, and Tool Hang.
Lift requires the robot to grasp and lift an object; Can requires the robot to grasp a cylindrical object and place it at a target location; Square involves precisely inserting a square nut onto a peg; and Tool Hang is a long-horizon multi-stage assembly task where the robot constructs a stand and hangs a tool.
For LIBERO, we evaluate on five tasks following the experimental protocol of prior work~\citep{dong2026fastervalueguidedsamplingfast}.

\textbf{Baselines. }
We compare against both state-of-the-art model-free algorithms and model-based algorithms. We refer to \Cref{ap:comparisons} for details on algorithmic comparisons. 

\subsection{How Does \ours{} Perform Compared to State-of-the-Art Model-Free and Model-Based RL Methods?}
\begin{figure}[t]{
  \centering
  \includegraphics[width=\linewidth]{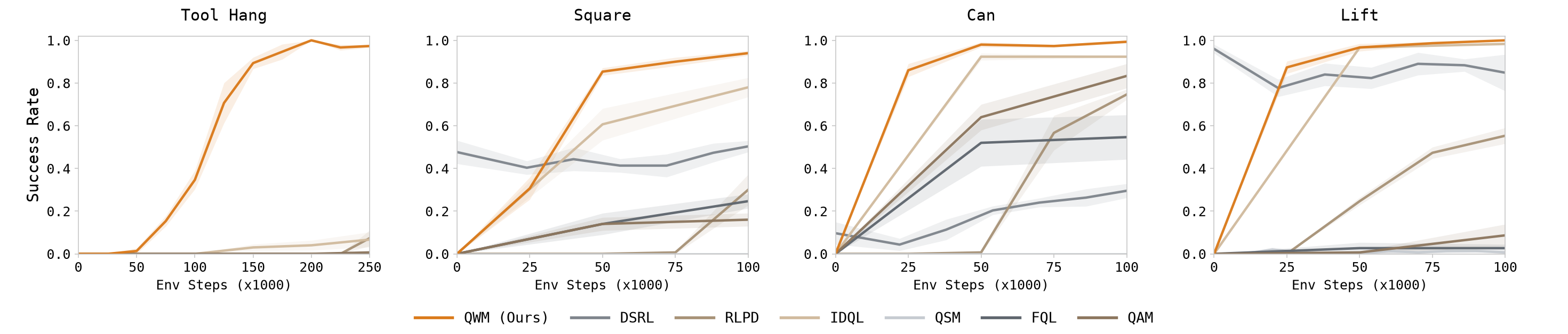}
  \textbf{Compared with model-free baselines.}
  Success rates of \ours{} and model-free baselines in the online settings.
  \ours{} outperforms strong model-free baselines in sample efficiency.
  }
  \label{fig:model_free_baselines}
\end{figure}

\begin{figure}[t]
  \centering
  \includegraphics[width=\linewidth]{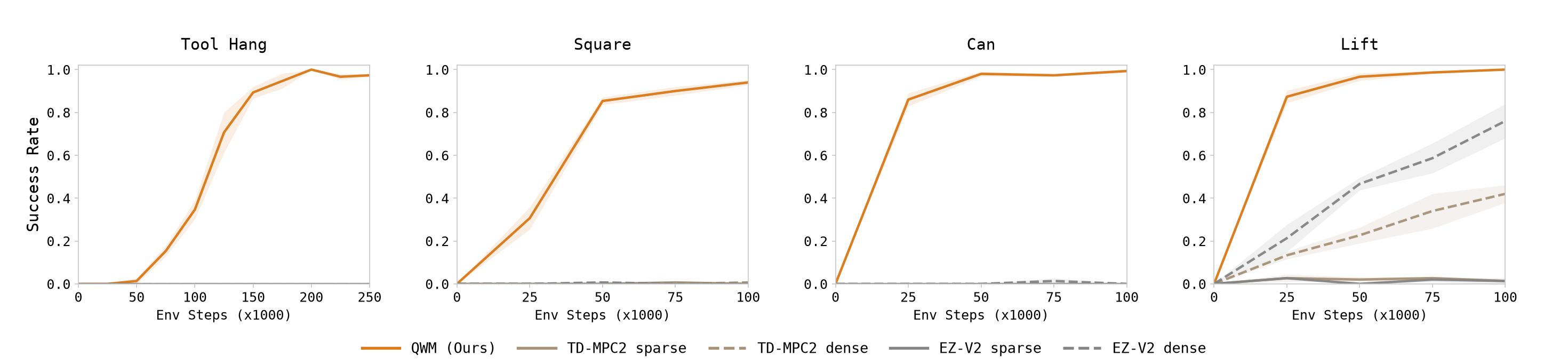}
  \caption{\footnotesize
  \textbf{Compared with model-based RL methods.}
  Success rates of \ours{}, TD-MPC2, and EZ-V2 across Lift, Can, Square, and Tool Hang under sparse and dense reward settings.
  \ours{} achieves strong performance across challenging manipulation tasks.
  }
  \label{fig:model_based_rl}
\end{figure}

\textbf{Model-free.} We compare against strong model-free RL baselines in \Cref{fig:model_free_baselines}, all of which leverage online interaction to learn Q-functions. \ours{} achieves the strongest performance across all tasks. RLPD, DSRL, QSM, QAM, and FQL each sample a single action from the policy for execution in the environment. While this action is intended to approximately maximize the Q-function, optimization delays and errors prevent it from being truly optimal. \ours{} instead leverages the world model's ability to predict future outcomes, performing test-time search over candidate actions and selecting the ones with the best predicted future returns. While IDQL selects actions by sampling multiple candidates and selecting the one with the highest Q-value, it considers only the immediate next action rather than a sequence of future actions, the latter carrying substantially more signal about long-term performance. These results demonstrate that the test-time search in \ours{} provides benefits complementary to Q-learning, enabling better action selection without substituting online interactions.

\textbf{Model-based.} We further compare \ours{} with state-of-the-art model-based RL methods TD-MPC2 and EfficientZero V2. We include both sparse and dense reward variants for the model-based algorithms, as many of these methods were designed for dense rewards, and struggles to learn as well in sparse reward settings. We present the results in \Cref{fig:model_based_rl}. \ours{}
achieves consistently stronger performance across the evaluated manipulation
tasks, while TD-MPC2 and EfficientZero V2 only obtain non-zero success on Lift for the number of training steps reported. Different from conventional model-based RL methods that incorporate model
predictions into policy optimization or learning targets, \ours{} uses the
world model only at test-time search while keeping policy and value learning
grounded on real environment transitions. This design allows \ours{} to
leverage future predictions for improved action selection without relying on
model-generated trajectories during learning, providing an effective way to
combine the predictive capability of world models with online Q-learning while avoiding compounding bias from the world model. 

\subsection{How Does \ours{} Perform Compared to the Base Method It Is Implemented on Top of?}

\begin{figure}[t]
  \centering
  \includegraphics[width=\linewidth]{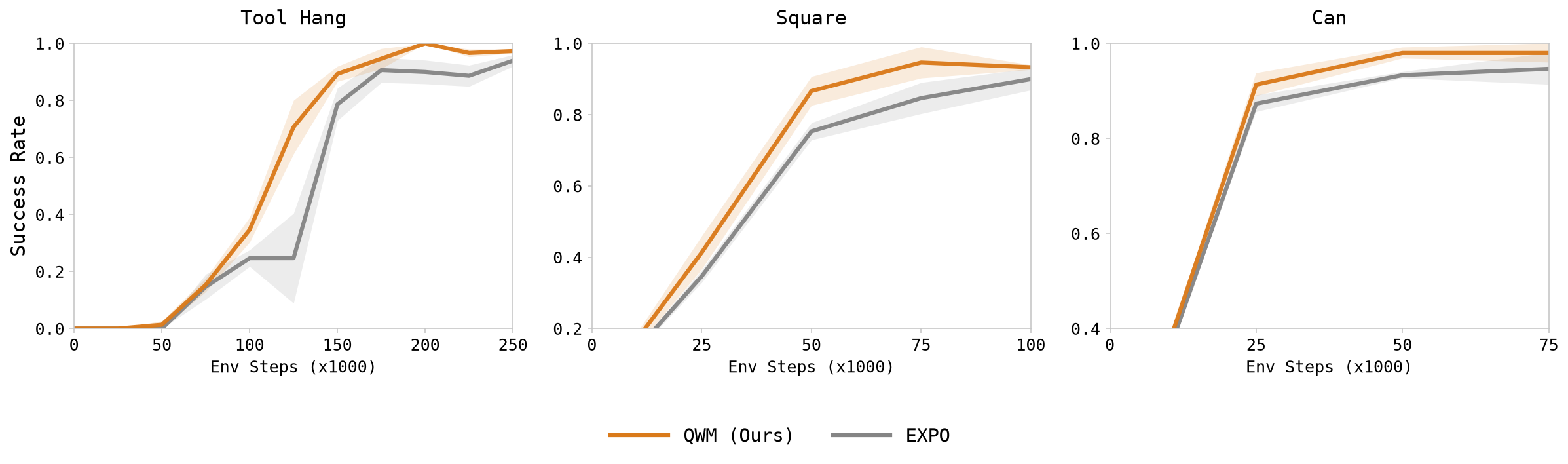}
  \caption{\footnotesize
  \textbf{Effect of world-model tree search on EXPO.}
  Online success rates of EXPO and \ours{} on Tool Hang, Square, and Can.
  Tree search consistently improves sample efficiency, with larger gains on Tool Hang and Square.
  }
  \label{fig:tree_search_ablation}
\end{figure}

\begin{figure}[t]
  \centering
  \includegraphics[width=\linewidth]{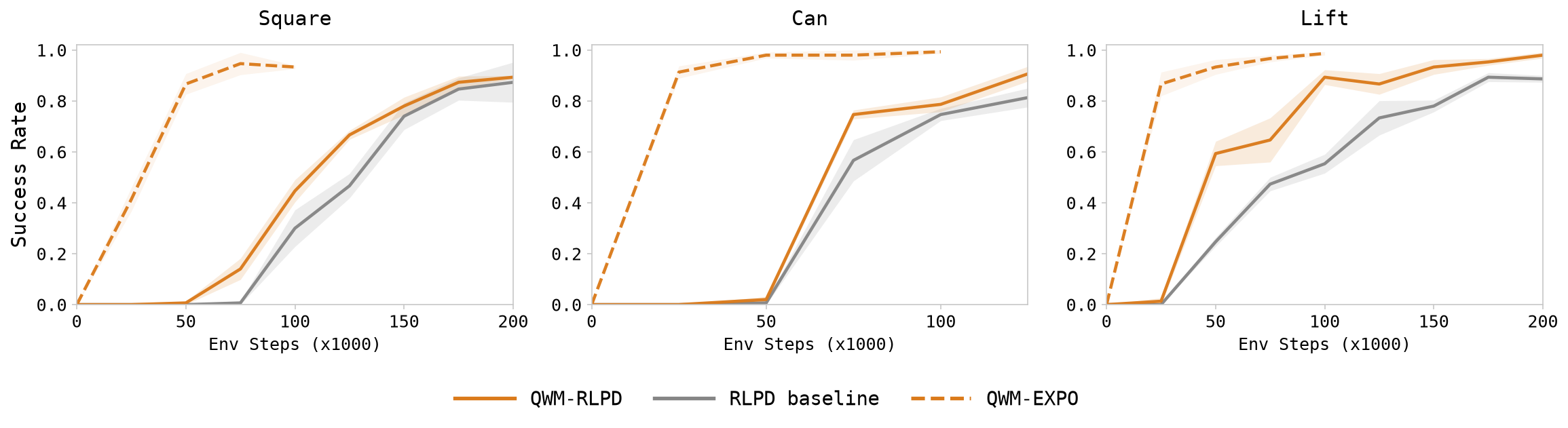}
  \caption{\footnotesize
  \textbf{Effect of world-model tree search on RLPD.}
  Online success rates of RLPD, RLPD with tree search, and \ours{} on Square, Can, and Lift.
  Tree search improves the sample efficiency of RLPD across all three tasks, while \ours{} built on EXPO remains more sample-efficient overall.
  }
  \label{fig:rlpd_tree_search}
\end{figure}

In this section, we evaluate the performance of \ours{} compared to the base algorithm it is implemented on. We implement \ours{} on top of EXPO and RLPD for the paper as representative sample efficient Q-learning algorithms, but \ours{} can be implemented on other model-free algorithms as well. We start by comparing \ours{}
against EXPO in
\Cref{fig:tree_search_ablation}. \ours{} consistently
improves learning efficiency across Tool Hang, Square, and Can, with
particularly pronounced gains on harder tasks such as Tool Hang. Comparing with RLPD as the base algorithm, as shown in Figure 6, using \ours{}
with RLPD consistently improves learning compared to RLPD alone with particularly clear gains in sample efficiency. These findings
show that \ours{} provides consistent benefits across different
underlying RL algorithms. At the same time, \ours{} will benefit from a more sample efficient base algorithm as shown by the results of \ours{} on top of EXPO compared to \ours{} on top of RLPD. 
\begin{figure}[t]
  \centering
  \includegraphics[width=\linewidth]{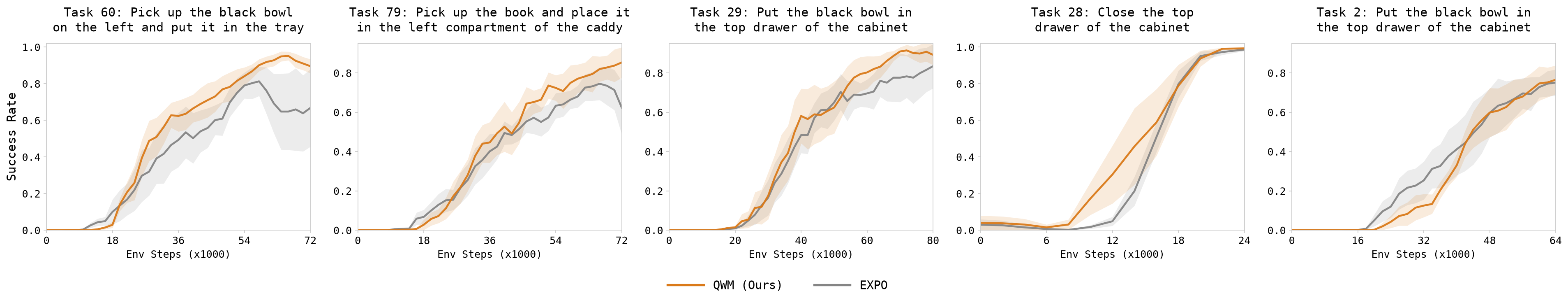}
  \caption{\footnotesize
  \textbf{Pixel-based evaluation on LIBERO.}
  Online success rates of \ours{} and EXPO across five LIBERO tasks.
  \ours{} achieves clear gains on Tasks 60 and 79, improves learning speed on Task 28, and reaches comparable or stronger final performance across the remaining tasks.
  }
  \label{fig:pixel_results}
\end{figure}

\subsection{Does \ours{} Scale to a Pixel-Based Setting?}
We further evaluate whether the performance gains of \ours{} extend to high-dimensional visual observations. Due to computational constraints, we use the world model only for sampling during online RL data collection, not for evaluation. As shown in \Cref{fig:pixel_results}, \ours{} achieves a clear overall improvement over EXPO. The gains are most evident on Tasks 60, 79, and 29, where \ours{} learns faster and reaches stronger late-stage performance. On Task 28, both methods eventually attain near-perfect success, but \ours{} reaches high performance earlier, demonstrating a clear improvement in sample efficiency. As shown in the ablations in \Cref{fig:tree_search_ablation}, restricting the world model to sampling-only (without evaluation) reduces performance, suggesting the visual results can be even stronger given sufficient compute. Nonetheless, even in this sampling-only setting, \ours{} consistently improves over the base algorithm, showing that its benefits extend naturally to high-dimensional visual inputs, which are substantially harder to learn a dynamics model from.

\begin{figure}[t]
  \centering
  \includegraphics[width=\linewidth]{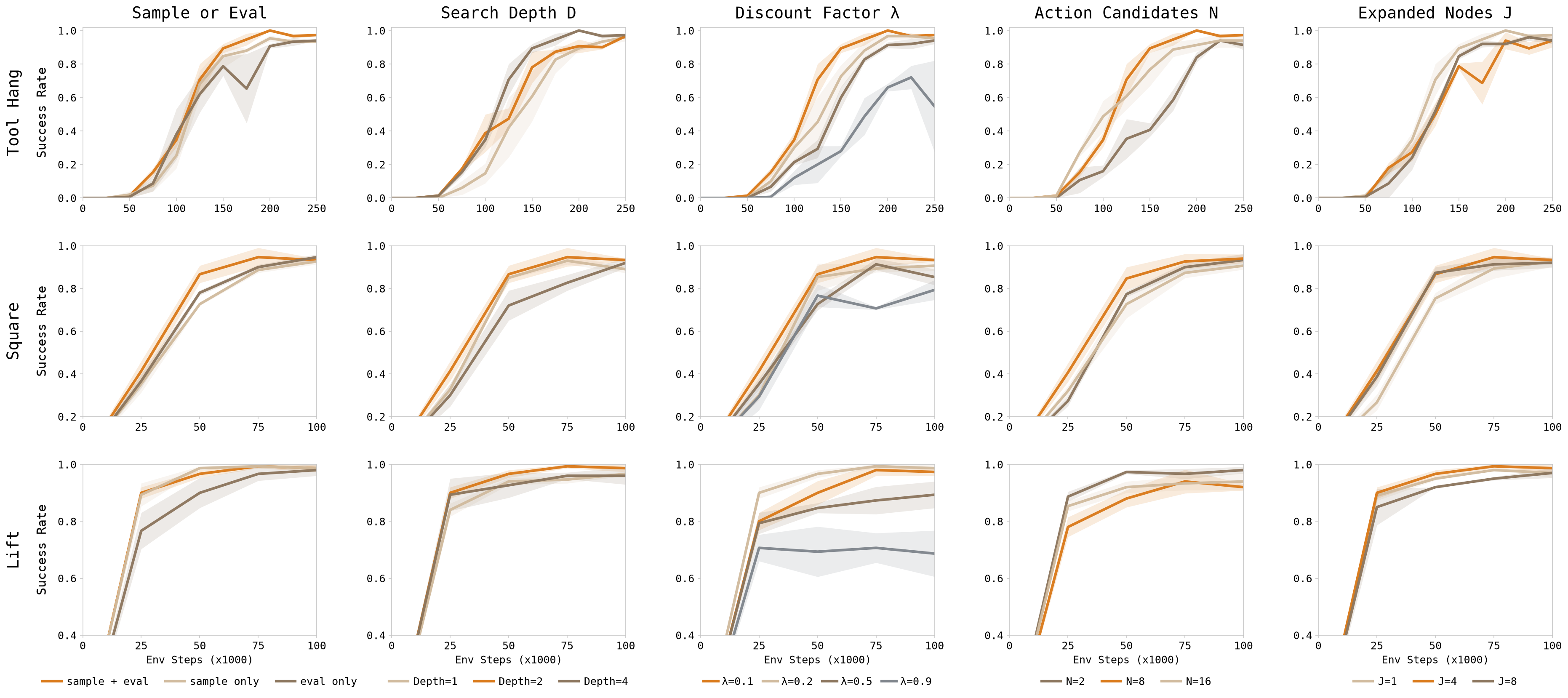}
  \caption{\footnotesize
  \textbf{Ablation of world-model lookahead on state-based tasks.}
  We compare applying world-model-guided action selection during online
  sampling, evaluation, or both stages. The \textit{sampling + evaluation}
  variant achieves the most consistent improvement across tasks, showing the
  complementary benefits of improving online experience collection and
  test-time action selection.
  }
  \label{fig:state_ablation}
\end{figure}

\subsection{What Components of \ours{} Are Most Important for Performance?}

To better understand the significance of different pieces of \ours{}, we ablate over three key components: (1) the importance of using \ours{} during online RL sampling, evaluation, or both, (2) the hyperparameters for constructing the tree, and (3) the number of expanded nodes during search to constrain the tree size. We present additional experiments comparing against performing search with only a state value function \Cref{ap:additional}.

\textbf{Sampling versus evaluation.}
We first ablate whether \ours{} is applied during online RL data collection, at evaluation time, or in both. The \textit{sampling only} variant performs test-time search during online data collection but disables it at evaluation, while the \textit{evaluation only} variant applies tree search only at evaluation time. The full \textit{sampling + evaluation} variant uses test-time search in both. As shown in \Cref{fig:state_ablation}, applying test-time search at both stages provides the strongest and most consistent learning efficiency. The full variant improves faster and more smoothly and reaches high success rates earlier. Sampling-time search affects which transitions are collected and added to the replay buffer, allowing subsequent policy and critic updates to benefit from higher-value online experience. Evaluation-time search, in contrast, directly improves action selection without modifying the learned policy or training data. Performing test-time search at both stages therefore combines improved online experience collection with stronger execution-time action selection, which gives more consistent advantage over either single-stage variant.

\textbf{Tree-search hyperparameters.}
We next ablate key hyperparameters for constructing the search tree: search depth $D$, recursive discount factor $\lambda$, and the number of action candidates $N$. In \Cref{fig:state_ablation}, we see that search depth has a clearer effect on learning. Depth $2$ generally provides the strongest performance gain in our experiments, whereas depth $1$ improves more slowly as a shallow search may not capture enough future consequence to reliably distinguish candidate actions. For tasks requiring longer term planning, higher depths can be beneficial, though making the depth too high could expose the search to additional world-model prediction error and uncertainty. Consequently, a moderate search depth provides a favorable trade-off between incorporating future information and maintaining reliable predictions.

The value aggregation discount factor $\lambda$, which controls how strongly future tree values contribute to the recursive value aggregation in \Cref{eq:recursive-value} is important for performance and setting $\lambda$ too large or too small leads to lower performance, and relatively small to moderate values, particularly $\lambda=0.2$ for our settings, provide the strongest learning efficiency. A larger $\lambda$ places greater weight on values propagated from deeper imagined states, making the value aggregation more sensitive to accumulated world-model and value-estimation errors, while an overly small $\lambda$ limits the contribution of a deeper search. Overall, the results favor large weighting of short-term estimates with a moderately small future-value weighting.

We next examine the effect of the number of action candidates 
$N$ considered at each node of the search tree. Increasing 
$N$ allows the search to evaluate a broader set of possible actions before committing to the one that maximizes the recursive value estimate, which in principle should improve the quality of the selected action by reducing the chance that a good candidate is overlooked. Empirically, we find that the number of actions $N$ that performs the best is dependent on the environment. In general, performance improves as 
$N$ increases from very small values, since too few candidates can fail to include actions that meaningfully diverge from one another, limiting the benefit of search altogether. However, the gains from increasing 
$N$ diminish beyond a moderate number of candidates, and excessively large 
$N$ can even hurt performance, likely because evaluating many candidates amplifies the influence of world-model and value-estimation errors across a larger number of imagined rollouts, while also increasing computational cost with limited additional benefit. We find that a moderate number of action candidates offers the best trade-off between exploration of the action space and robustness to compounding prediction errors, and we adopt this setting as default in our main experiments.

\textbf{Number of expanded nodes. }
Lastly, we ablate over the number of expanded nodes in our search heuristic. In \Cref{fig:state_ablation}, we observe that performance is relatively insensitive to the number of expanded nodes $J$, as the heuristic accounts for which paths have the largest value to decide which nodes to keep expanding and which to prune. While this greedy approach does not guarantee finding the highest value paths of the full tree, it provides a good approximation. This suggests that retaining a small number of high-value branches is already sufficient to capture useful candidate futures.

%% file: discussion.tex
\section{Discussion} \label{sec:discussion} 

In this paper, we present \ours{}, a framework for leveraging world models  via test-time search over actions on top of Q-learning to improve performance. Rather than using the world model to optimize the policy directly, as in conventional model-based RL, we use it purely at test time both during online rollouts and at evaluation. We show that \ours{} outperforms strong model-free baselines, while avoiding the compounding model bias that afflicts methods which optimize policies primarily inside a learned world model. Despite these results, \ours{} has limitations. The tree search introduces non-trivial computational overhead at rollout and evaluation time relative to a standard policy forward pass, which may be prohibitive under tight latency requirements. Additionally, \ours{} depends on learning a world model, which is an expensive and often challenging to train; while grounding the policy and Q-function in real data prevents bias from the model from compounding, reducing search overhead and obtaining high quality world models remain important future directions.

\section{Acknowledgments}
This work was supported in part by an NSF CAREER award, NSF \#1941722 the RAI Institute, ONR grant N00014-22-1-2293, and ONR grant N00014-22-1-2621. This work used the Delta system at the National Center for Supercomputing Applications [award OAC 2005572] through allocation CIS260152 from the Advanced Cyberinfrastructure Coordination Ecosystem: Services \& Support (ACCESS) program, which is supported by U.S. National Science Foundation grants \#2138259, \#2138286, \#2138307, \#2137603, and \#2138296.